\documentclass[letterpaper, 10 pt, conference]{ieeeconf}  % Comment this line out if you need a4paper
\usepackage{amssymb}                                         
\usepackage[linesnumbered,lined,ruled,vlined,commentsnumbered]{algorithm2e}
\usepackage{graphicx}
\usepackage{caption}
\usepackage{subcaption}

\usepackage{ifthen,version}
\newboolean{include-notes}
\usepackage[usenames,dvipsnames]{color}
\setboolean{include-notes}{true}
\newcommand{\rknote}[1]{\ifthenelse{\boolean{include-notes}}%
 {\textcolor{Dandelion}{\textbf{RK: #1}}}{}}
\newcommand{\ernote}[1]{\ifthenelse{\boolean{include-notes}}%
 {\textcolor{cyan}{\textbf{#1}}}{}}

\IEEEoverridecommandlockouts                              % This command is only needed if 
\makeatletter
\let\NAT@parse\undefined
\makeatother
\usepackage[numbers, sort&compress]{natbib}
\usepackage[bookmarks=true]{hyperref}
\usepackage[nameinlink,capitalize]{cleveref}

\hypersetup{letterpaper,bookmarksopen,bookmarksnumbered,
pdfpagemode=UseOutlines,
colorlinks=true,
linkcolor=blue,
anchorcolor=blue,
citecolor=blue,
filecolor=blue,
menucolor=blue,
urlcolor=blue
}

\title{\LARGE \bf An Information-Theoretic Approach to Persistent Environment Monitoring Through Low Rank Model Based Planning and Prediction
}

\author{Elizabeth A. Ricci$^{1}$, Madeleine Udell$^{2}$, and Ross A. Knepper$^{1}$% <-this % stops a space
\thanks{*This work was supported by the Air Force Office of Scientific
Research under award number FA9550-17-1-0109.}% <-this % stops a space
\thanks{$^{1}$These authors are with the Department of Computer Science,
        Cornell University, Ithaca, NY, USA
        {\tt\small \{ericci,rak\}@cs.cornell.edu}}%
\thanks{$^{2}$The author is with the Department of Operations Research and Information Engineering,
        Cornell University, Ithaca, NY, USA
        {\tt\small udell@cornell.edu}}%        
}

\begin{document}

\maketitle
\thispagestyle{empty}
\pagestyle{empty}

%%%%%%%%%%%%%%%%%%%%%%%%%%%%%%%%%%%%%%%%%%%%%%%%%%%%%%%%%%%%%%%%%%%%%%%%%%%%%%%%
\begin{abstract}

Robots can be used to collect environmental data in regions that are difficult for humans to traverse. However, limitations remain in the size of region that a robot can directly observe per unit time. We introduce a method for selecting a limited number of observation points in a large region, from which we can predict the state of unobserved points in the region. We combine a low rank model of a target attribute with an information-maximizing path planner to predict the state of the attribute throughout a region. Our approach is agnostic to the choice of target attribute and robot monitoring platform. We evaluate our method in simulation on two real-world environment datasets, each containing observations from one to two million possible sampling locations. We compare against a random sampler and four variations of a baseline sampler from the ecology literature. Our method outperforms the baselines in terms of average Fisher information gain per samples taken and performs comparably for average reconstruction error in most trials.

\end{abstract}

\newcommand{\region}{\ensuremath{\mathcal{R}}}
\newcommand{\property}{\ensuremath{p}}
\newcommand{\data}{\ensuremath{D}}
\newcommand{\robot}{\ensuremath{r}}
\newcommand{\start}{\ensuremath{s}}
\newcommand{\budget}{\ensuremath{b}}
\newcommand{\cost}{\ensuremath{C}}
\newcommand{\Path}{\ensuremath{P}}
\newcommand{\observations}{\ensuremath{\omega}}

\section{Introduction}\label{sec:introduction}

Ecologists and other researchers study environmental features by analyzing data regularly collected over an extended period of time~\citep{smith2011persistent2,wang2010review,ma2016information,ma2018data,langford1993unmanned}. However, exhaustively collecting these data is expensive, sometimes physically infeasible, and each individual collection must be completed within a given timeframe. Robots can be used to automate this process and reduce the danger to humans, but limitations on robot actuation, sensing, and battery life prevent robots from observing every point in a large geographic region within the required timeframe. 

We mitigate these obstacles to autonomous persistent environment monitoring through a method for sampling a cost-budgeted set of observation points in a large region. From these sampled observations, we can predict the state of an environmental feature for the entire region.
% Due to limitations on robot battery life, system memory constraints, and environmental factors (daylight, wind, extreme weather conditions, etc.) that limit robot motion and sensor accuracy, taking observations at every point a large geographical region is unrealistic. 
Our approach exploits local structural correlations in environmental measurements over a region to fit a low rank model. This model, in combination with an information-maximizing planner, allows us to select a set of observation points for a robot to sample from. This set is selected such that these samples minimize the amount of uncertainty about the value of the observed attribute throughout the region~\cite{shannon1948mathematical}. Further, this maximization is constrained by a user-provided cost function and budget.

In the remainder of this work, we describe the details of our low rank model and information-maximizing path planner, and present an evaluation of our approach on real-world environmental measurement datasets. These datasets are large (each contains one to two million possible sampling locations) and measure features of interest to the scientific community; this evaluation demonstrates the performance of our technique on a realistic persistent environment monitoring task.

 \begin{figure*}[t]
            \centering
         \includegraphics[width=0.95\textwidth]{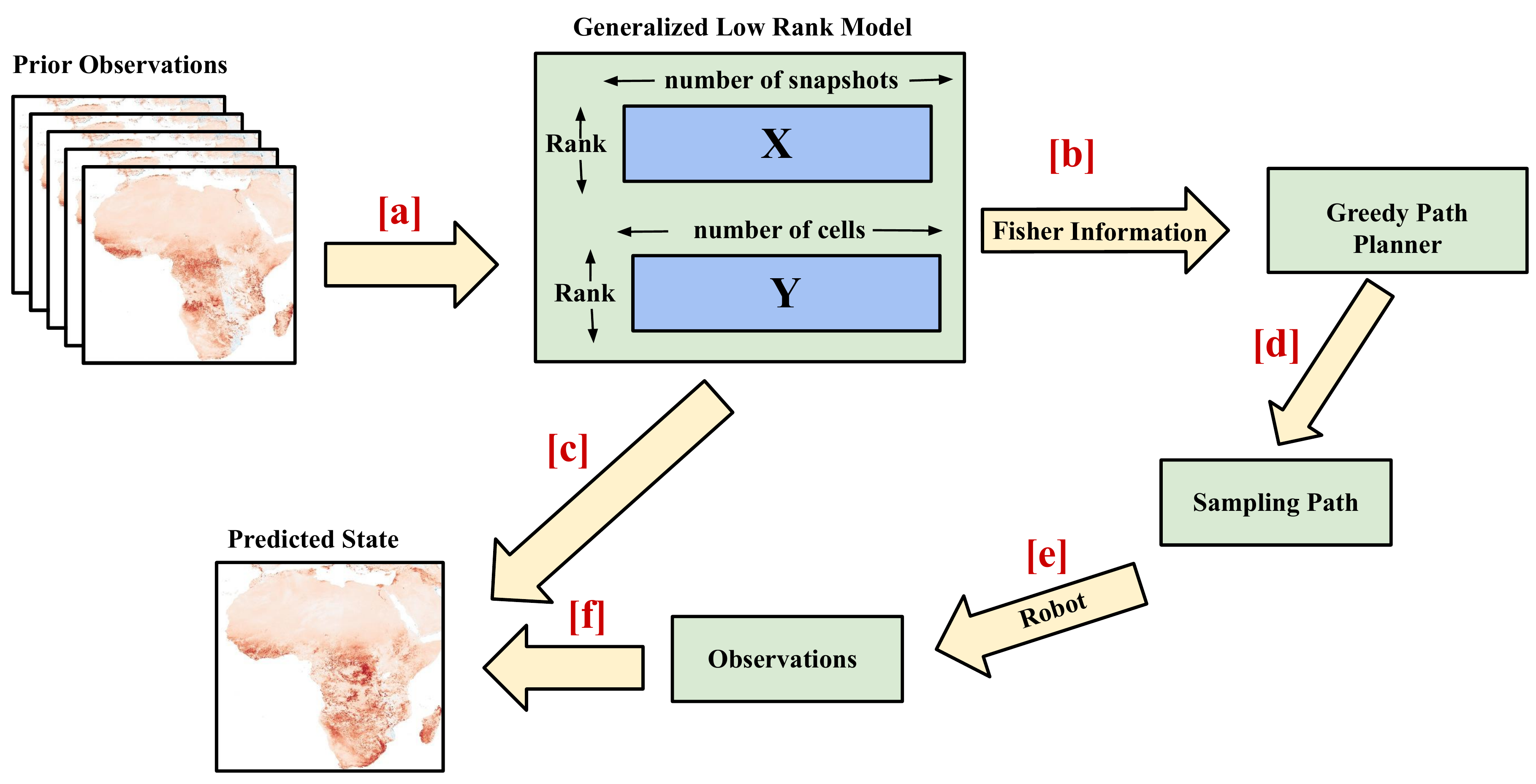}

            \caption{Prior observations are used to generate a generalized low rank model (arrow \textbf{[a]}). Fisher information computed from the GLRM (arrow \textbf{[b]}), is used by the path planner to form a sampling path (arrow \textbf{[d]}). The robot takes observations along the path (arrow \textbf{[e]}). The observations are used (arrow \textbf{[f]}), in conjunction with the GLRM (arrow \textbf{[c]}) to predict the states of the unobserved cells.}
   \label{fig:approach}
    \end{figure*}
    
\section{Related Work}\label{sec:relatedWork}

Persistent monitoring is used to collect long-term data for study of time-varying environmental attributes~\citep{wang2010review,ma2016information,smith2011persistent2, ma2018data, smith2011persistent}. This monitoring can be performed manually or autonomously, using robot platforms~\citep{ma2016information,smith2011persistent2,langford1993unmanned, ma2018data,smith2011persistent, yu2014correlated}. Many approaches to autonomous persistent monitoring use information-based planning to select data-collection locations within a large monitoring region~\citep{ma2016information, ma2018data, charrow2015information, low2011active}.

% Persistent monitoring~\citep{ma2016information,smith2011persistent2,langford1993unmanned, ma2018data,smith2011persistent, yu2014correlated} can use a robot to take samples and capture temporal variations within the region. Examples of attributes include: vegetation health~\citep{wang2010review}, ocean properties~\citep{ma2016information,smith2011persistent2, ma2018data}, hurricane predictors~\citep{langford1993unmanned}, survivor locations~\citep{singh2009nonmyopic}, and wireless signal strength~\citep{hollinger2014sampling}.  

Information-based path planning problems concern the maximization of an information measure, subject to constraints on robot sensing and motion. An information measure quantifies the reduction in uncertainty of a probabilistic value due to a particular observation~\cite{shannon1948mathematical}. Therefore, maximizing an information measure minimizes the global uncertainty of the value of the observed attribute. Greedy and sampling-based path planning algorithms are commonly used to solve this class of problems. Greedy solutions are performant because many information measures are submodular ~\citep{sharma2015greedy, hashemi2018randomized}; greedy algorithms are known to approximately maximize submodular functions to within $(1-\frac{1}{e})$ of optimal ~\citep{nemhauser1978analysis,ko1995exact, jorgensen2018team, ma2018data, nguyen2016real}. Classical sampling-based planners can also be adapted for use in the information-based planning; these planners are efficient and scale well to large problem instances \cite{hollinger2014sampling, kwak2008path, zhang2009information, lu2014information}. Our method uses the greedy approach introduced in \cite{zhang2016submodular}.

The information measure and set of constraints for a given information-based planning problem are specific to that problem's target application. Our work maximizes Fisher information~\cite{frieden_2004} subject to a maximum path length constraint; this is similar to problems from the literature \cite{hollinger2014sampling, leung2006planning, yu2014correlated, singh2009nonmyopic}.

% Our chosen problem formulation lends itself to high-level solutions where specific motion decisions are deferred to the controller. As a result, the methods used are fairly general and can be applied to a wide variety of robotic systems. 

Some approaches explicitly incorporate constraints such as obstacles, environment perturbations, and robot kinodynamics into the planning problem \cite{cai2009information, lu2014information, zhang2009information, ma2016information, nguyen2016real}. Our approach frames the planning problem as determining a set of observation locations and leaves trajectory generation for a specific robot platform to a lower level controller. This abstracts our approach from any specific robot platform and allows greater flexibility in execution. 

% Considering multiple constraints in the path planning process results in an increase in control, as the plans produced are in the form of a lower-level motion plan. This gives more performant solutions for application-specific planners, but requires more initial information and results in the loss of flexibility.

% Our approach only considers path length constraints during the planning process. This results in a path containing waypoints for a robotic system to sample from, solutions to the aforementioned problems are in the form of a lower-level motion plan. This gain in control allows for more performant solutions for application-specific planners, but requires more initial information and results in the loss of flexibility.

% In the information-based planning domain, we want to minimize the uncertainty of the state of some object(s). Uncertainty is typically quantified using a form of entropy or information gain. These information functions can be computed using prior information about the state of the object(s) considered.

In this work we use Fisher information, computed from a generalized low rank model of the data, to measure the reduction of uncertainty as a result of taking samples. If we assume the data are sampled from a Gaussian process, then the Fisher Information is equal to the log determinant of the inverse of the covariance matrix~\cite{frieden_2004}. The Gaussian process assumption is similar to methods used in prior information-based path planning \cite{charrow2015information, low2009information, ma2018data, ma2016information, low2011active}. 

% In cases where the data is not normally distributed, Bayesian models can be used to model information containing many variables. This is seen in work where the information function is represented in terms of a non-Gaussian sensor objective \cite{zhang2009information, lu2014information,cai2009information}. In the sensor based approach, the Bayesian model encodes the probability mass function of sensor readings. As a result, after taking some sensor readings, the conditional entropy measures the amount of uncertainty of the value of all other sensor readings. This differs from our approach, which considers the Fisher information \ernote{add more info?} of the states of the sampling points. With this representation, the conditional entropy represents the amount of uncertainty about the unseen sampling points after a number of sampling points are observed. \ernote{FISHER INFORMATION} In our approach we can more clearly consider the mutual information between a set of targets or sampling points due to our assumption that a target can be completely observed at some sampling location.

There is work on information-directed sampling for low rank matrix completion for an application in automated machine learning, but the method does not consider physical constraints~\citep{yang2018oboe}.

\section{Problem Formulation}
Let \region{} be a grid-discretized two-dimensional region composed of $L$ cells, or sampling locations. Let \property{} be a continuous-valued environmental attribute that can be observed at every cell in \region{}. We assume the value of \property{} at every cell in \region{} is time-varying, but changes relatively slowly with respect to the sampling frequency. For instance, the hue of foliage over the northeast region of the Americas in the Fall can be considered static over the span of a few hours, but is dynamic over the span of a few days. Therefore, for this domain, observations taken within a few hours of each other are considered to be sampled at the same time. 

We have a dataset containing a set of snapshots of \property{} in \region{} for $T$ different times. Each snapshot contains observations of the value of \property{} in a subset of \region{}. 
Let $\cost(P): \mathcal{P}(\region) \to \mathbb{R}$, where $P \subseteq \region$, be a cost function from a subset of cells in \region{} to a real value. This function computes the cost of traveling between a set of cells. Let $\budget\in \mathbb{R}$ be a budget corresponding with \cost{}.

We consider the problem of choosing a set of sampling locations, $P \subseteq \region$ such that: (1) $\cost(P) \leq \budget$, the cost of $P$ is within the budget, and (2) the uncertainty about the state of \property{} over \region{} is minimized after observations are taken for all cells in $P$. We maximize Fisher information as a proxy for minimizing uncertainty. Additionally, we want to predict the current state of the unobserved cells in \region{}, given, \observations{}, a recently-sampled set of observations of \property{} in \region.

\section{Approach}\label{sec:approach}

The prior observations are represented as a data matrix and used to fit a generalized low rank model (GLRM)~\cite{udell2016generalized} composed of matrices $X$ and $Y$, as shown in \cref{fig:approach}, arrow \textbf{[a]}. The GLRM captures the structure of \property{} both spatially and temporally and is used to compute Fisher information for sets of cells in \region{}.

The greedy path planner uses Fisher information (\cref{fig:approach}, arrow \textbf{[b]}), a cost function, and a cost budget to choose a set of sampling points in \region{}, starting an initial cell (\cref{fig:approach}, arrow \textbf{[d]}). The Fisher information of the sampling points is within $\frac{1}{2}(1-e^{-1})$ of optimal, given the constraints, and that the cost of visiting each point, per the cost function, is within the provided budget. The robot makes observations along this path (\cref{fig:approach}, arrow \textbf{[e]}) to get \observations{}, a set of observations.

Finally, \observations{} and the GLRM (\cref{fig:approach}, arrows \textbf{[f]} and \textbf{[c]}) are used to predict the states of the unobserved cells in \region{}.

\subsection{Generalized Low Rank Model}\label{sec:glrm}

We represent the state of the attribute \property{} over \region{} using a generalized low rank model (GLRM). A GLRM is a vector model that represents large datasets with missing and noisy data as a pair of comparatively small, low rank matrices \cite{udell2016generalized}. The low rank nature of this model means that it can predict unobserved values given a small number of accurate observations. In addition, Fisher information, representing the information of sets of cells as represented by the GLRM, can be computed using a generalized low rank model~\cite{yang2018oboe}.

Assume the environmental variable \property{} can be observed at $L$ locations within \region{} and observations are taken at $T$ different times. Prior observations of \property{} over \region{} are given as a dataset. We collect these prior observations in a data matrix, $D \in \mathbb{R}^{T \times L}$, where the value of $D_{i,j}$ is the value of \property{} observed at cell $i \in \region$ at the $j^{th}$ time. Unobserved values over the $T$ times and $L$ cells are set to arbitrary values in $D$; their values do not affect the results. A projection function $P_\Omega$ is used to ignore the entries of $D$ corresponding to missing or unobserved data. Let $\Omega \subseteq \{1,\ldots,T\} \times \{1, \ldots, L\}$ be the set of (time, location) tuples representing entries in $D$ that contain observations. Let $P_\Omega$ be the projection operator that ignores the unobserved values in $D$.

 We compute $X$ and $Y$ by minimizing: \[\quad \| P_\Omega( X^{\mathsf{T}}Y - D ) \|_F^2\]
with variables $X \in \mathbb{R}^{k \times T}$ and $Y \in \mathbb{R}^{k \times L}$, where $k\in \mathbb{Z}$ is the rank of the model. Intuitively, the columns of $Y$ are basis vectors for a latent space that encodes the spatial structure of \property{}; each column corresponds to one location in \region{}. Similarly, the columns of $X$ are basis vectors for a latent space that encodes the temporal structure of \property{}. The product $X^{\mathsf{T}}Y$ is a minimal-error representation of the original data matrix, including predictions for unobserved points.

\subsection{Path Planner}\label{sec:planner}

We use a greedy path planner to choose the set of sampling locations, given an information measure, and cost function with a corresponding cost budget. The specific planning algorithm is introduced by Zhang and Vorobeychik and maximizes a submodular quantity while respecting a cost budget~\cite{zhang2016submodular}. This planner requires a submodular function $\mathcal{I}: \mathcal{P}(\region) \to \mathbb{R}$ to maximize and an $\alpha$-submodular approximate cost function $\cost:\mathcal{P}(\region) \to \mathbb{R}$ with a corresponding budget value. Each of these functions take a set of cells and returns a real number. The specific cost and submodular functions chosen for the monitor application are discussed in \cref{sec:cost,sec:submod}. 

The planner selects a set of cells $\Path{} \subseteq \region$ from a given initial cell, \start{}. The set $\Path$ is initialized to contain the initial position, \start{}, and the $k-1$ closest cells to \start{}, where $k$ is the rank of the GLRM from \cref{sec:glrm}. The extra cells are added because Fisher information is a meaningful quantity only if the number cells is be greater than or equal to the GLRM's rank. 

At each iteration of the algorithm, the cell that maximizes the ratio of change in information to the change in cost is added to the set of sampling locations. Specifically, for each cell $x \in \region$ , a set $P' = \Path \cup \{x\}$ is formed. If $\cost(P')$ is less than or equal to the budget, we compute: 
\[\Delta_x = \frac{\mathcal{I}(P') - \mathcal{I}(\Path)}{\cost(P') - \cost(\Path)}.\]

Otherwise, if  $\cost(P')$ is greater than the budget, $x$ is not a valid option and $\Delta_x$ is set to zero. After iterating through each cell in the environment, a cell with the maximum value of $\Delta_x$ is added to the path $P$. This process is repeated until the maximum value of $\Delta_x$ is zero, that is, there is no cell to add to $P$ that would respect the cost budget. 

The path planner returns $P$, the set of cells in \region{} where the robot must take observations of \property{}. The cells in $P$ are treated as waypoints, that is, the actual trajectory of the robot should pass through each cell in $P$, but the robot's trajectory is left to a lower level controller. 

We use a greedy planning approach because Fisher information is submodular. Greedy approaches are computationally fast, simple to implement, and are known to approximately maximize submodular quantities. Zhang and Vorobeychik \cite{zhang2016submodular} prove that the value of the submodular function of the cells selected is within $\frac{1}{2}(1-e^{-1})$ times of the maximum value of the submodular function possible subject to the the cost constraints.

\subsection{Cost Function}\label{sec:cost}

The cost function represents the limit on how far the robot can travel. We use a shortest path cost function and the nearest neighbor algorithm as the $\alpha$-submodular approximate cost function, required by the planner. The nearest neighbor algorithm is a greedy-approximation for shortest path problems~\citep{golden1980approximate}.

 Starting with a given point, the algorithm adds points to the path based on proximity to the current point. That is, at each time-step, the next point to be traveled to is the one that is closest to the current point. Given a set of points $P=\{v_1, \dots, v_n\},$ the nearest neighbor algorithm returns a travel ordering of these points, starting at $v_1,$ the ordering is notated $P'=\{v'_1, \dots, v'_n\}.$ The cost of the path is:
\[
    \cost(V') = \sum_{i = 1}^{n-1} d(v'_i,v'_{i+1})\mbox{,}
\]
where $d(a,b)$ is the Euclidean distance between the points $a$ and $b$. 

For the planner's optimality guarantee to hold, as discussed in \cref{sec:planner}, the approximate cost function must be a $\psi(n)$-approximation, where $n$ is the number of points in the set and $\psi(n)$ is some constant real number. A cost function is a $\psi(n)$-approximation when it is at most $\psi(n)$ times greater than the true cost. In addition, the cost function should be $\alpha$-submodular. That is, when $x, A, B$ are minimized and $A \subset B$ the quotient \[\frac{cost(A \cup x)-cost(A)}{cost(B \cup x)-cost(B)}\] is equal to $\alpha$~\cite{zhang2016submodular}. The nearest neighbor cost function is a $\psi(x)$-approximation and $\alpha$-submodular~\citep{golden1980approximate}.

\subsection{Submodular Function}\label{sec:submod}

Fisher information is used to quantify information for the planning algorithm. Fisher information is a measure of relative information in a set of variables that can be computed directly from models, including the GLRM \cite{yang2018oboe,frieden_2004}. By observing the value of \property{} at the set of cell in \region{} that maximizes Fisher information, the robot will maximize information gain in expectation.

Formally, the Fisher information of a set of points, $V,$ is: 
\[
\mathcal{I}(V) = \log\det\Bigg(\sum_{p \in P} Y_pY_p^\mathsf{T}\Bigg)\mbox{,}
\]
where $Y$ is the matrix from the generalized low rank model as described in \cref{sec:glrm} and $Y_p$ is the column of $Y$ corresponding to the cell $p \in \region$. For the Fisher information to be a meaningful quantity, the number of points in the set must be greater than or equal to the number of rows in $Y$. Therefore we initialize the set of sampling points to include the $k - 1$ nearest cells to the initial cell, where $k$ is the rank of the GLRM. Fisher information is submodular because $\sum_{p \in P} Y_pY_p^\mathsf{T}$ is a positive definite matrix and therefore Fisher information compatible with the greedy planning algorithm~\cite{hashemi2018randomized}.

\subsection{Region Completion}\label{sec:prediction}

After the robot collects observations, the GLRM is used to predict the unobserved cells of the region. Region completion allows the robot to predict the full state of the region, without fully observing it.

Let $D,X,Y$ be as described in \cref{sec:glrm}, then the values of the unobserved cells are a linear combination of the columns of $Y$, let $x$ be the latent factor needed to determine the linear weights. 

We view the set of observations made as a new row $d \in \mathbb{R}^L$ of $D$ and a new column $x\in \mathbb{R}^k$ of $X$. Since the set of new observations do not cover the region, $d$ has missing entries. We define the subset of \region{} that was observed as $S \subseteq \{1,\ldots,L\}$. Define $Y_S \in \mathbb{R}^{k \times |S|}$ to collect those columns of $Y$ corresponding to indices in $S$. This allows the unobserved cells to be ignored in the reconstruction process. In order to estimate x, we minimize $x$ in the following: 

\[  \| P_S( Y^\mathsf{T} x - d ) \|^2 =\| Y_S^\mathsf{T} x - d_S \|^2 .\]

This can be reduced to: \[ (Y_S^\mathsf{T} Y_S)^\dagger Y_S^\mathsf{T} d_S,\]
where $A^\dagger$ represents the pseudoinverse of $A$. Note that the unobserved values are not included in the objective and therefore do not affect the solution.

Once $x$ is computed, the values of \property{} for the unobserved cells are predicted by $Y^\mathsf{T}x$.

% \]\[ (Y_S^\mathsf{T} Y_S)^\dagger Y_S^\mathsf{T} d_S =  (Y_S^\mathsf{T})^\dagger d_S,

% The robot's observations are stored in a row vector, $d_\textrm{path}$, and values for unobserved cells are set to zero.  We then compute a vector $x_\textrm{path}$ of dimension equal to the model rank by: \[
% x_\textrm{path} =  (Y^\texttt{T})^\dagger d_\textrm{path}
% \mbox{,}
% \]
% where $(Y^\texttt{T})^\dagger$ represents the pseudoinverse of $Y^\texttt{T}$. The vector $x_\textrm{path}$ represents the coefficients of the basis vectors in $Y$ used to calculate the observations contained in $d_\textrm{path}$. The missing values can be predicted by using the vector of constant values to scale the model's basis vectors:
% \[
%  d_\textrm{dense} =  Y^\texttt{T}x_\textrm{path}
% \mbox{.}
% \] That is, $d_\textrm{dense}$ contains the predicted values for each cell of the region, given the set of observations. 

% $$\mbox{argmin}_x \quad \| P_S( Y^T x - d ) \|^2 $$ $$ \mbox{argmin}_x \quad \| Y_S^T x - d_S ) \|^2$$ $$(Y_S^T Y_S)^\dagger Y_S^T d_S$$ $$(Y_S^T)^\dagger d_S$$

\section{Evaluation}\label{sec:evaluation}
\begin{figure*}[t]
        \centering
        \begin{subfigure}[b]{0.45\textwidth}
            \centering
            \includegraphics[width=\textwidth]{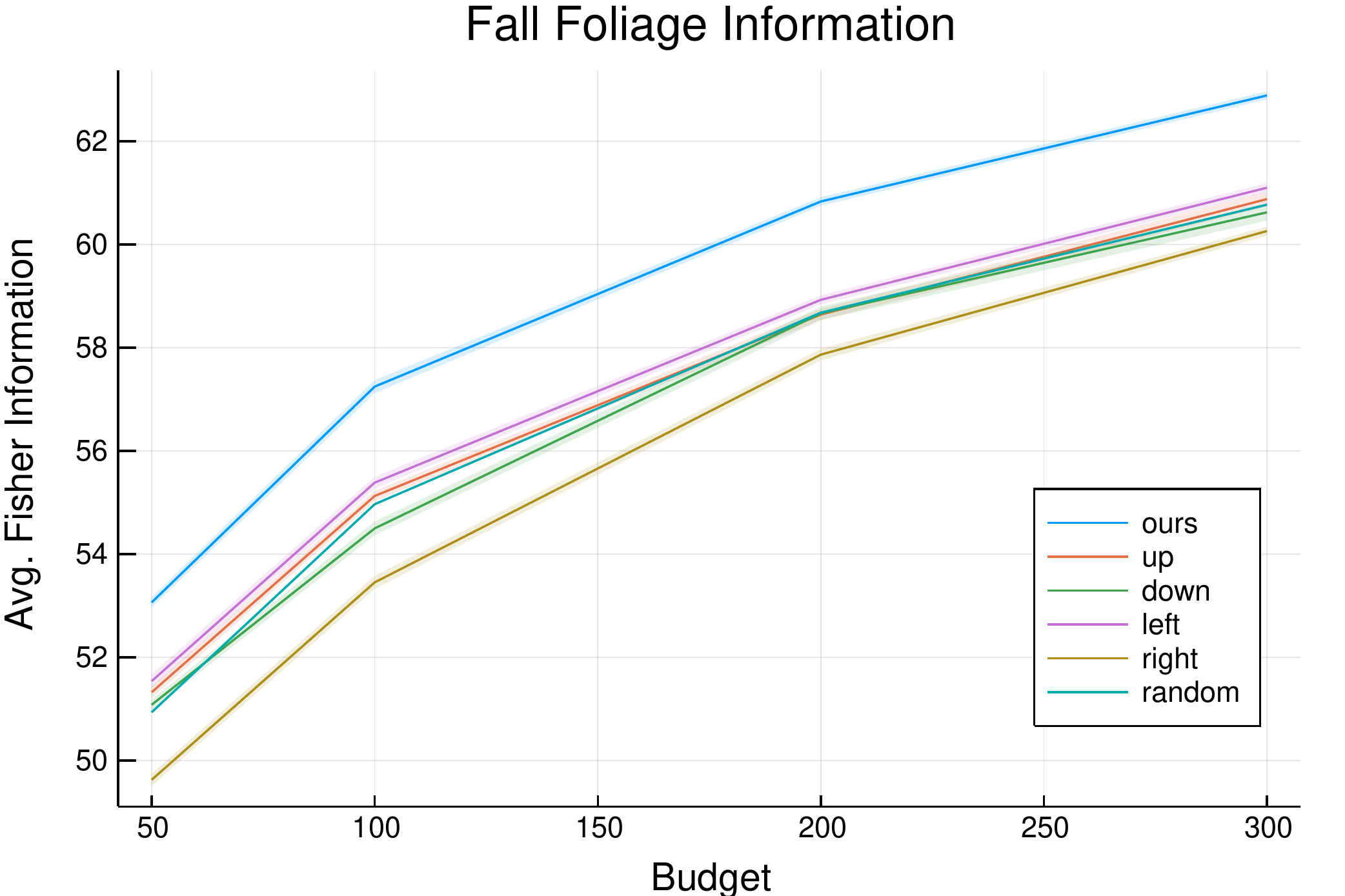}
               \caption{Fisher information for the Fall foliage dataset.}
            \label{fig:foliage_info}
        \end{subfigure}
        \hfill
        \begin{subfigure}[b]{0.45\textwidth}  
            \centering 
            \includegraphics[width=\textwidth]{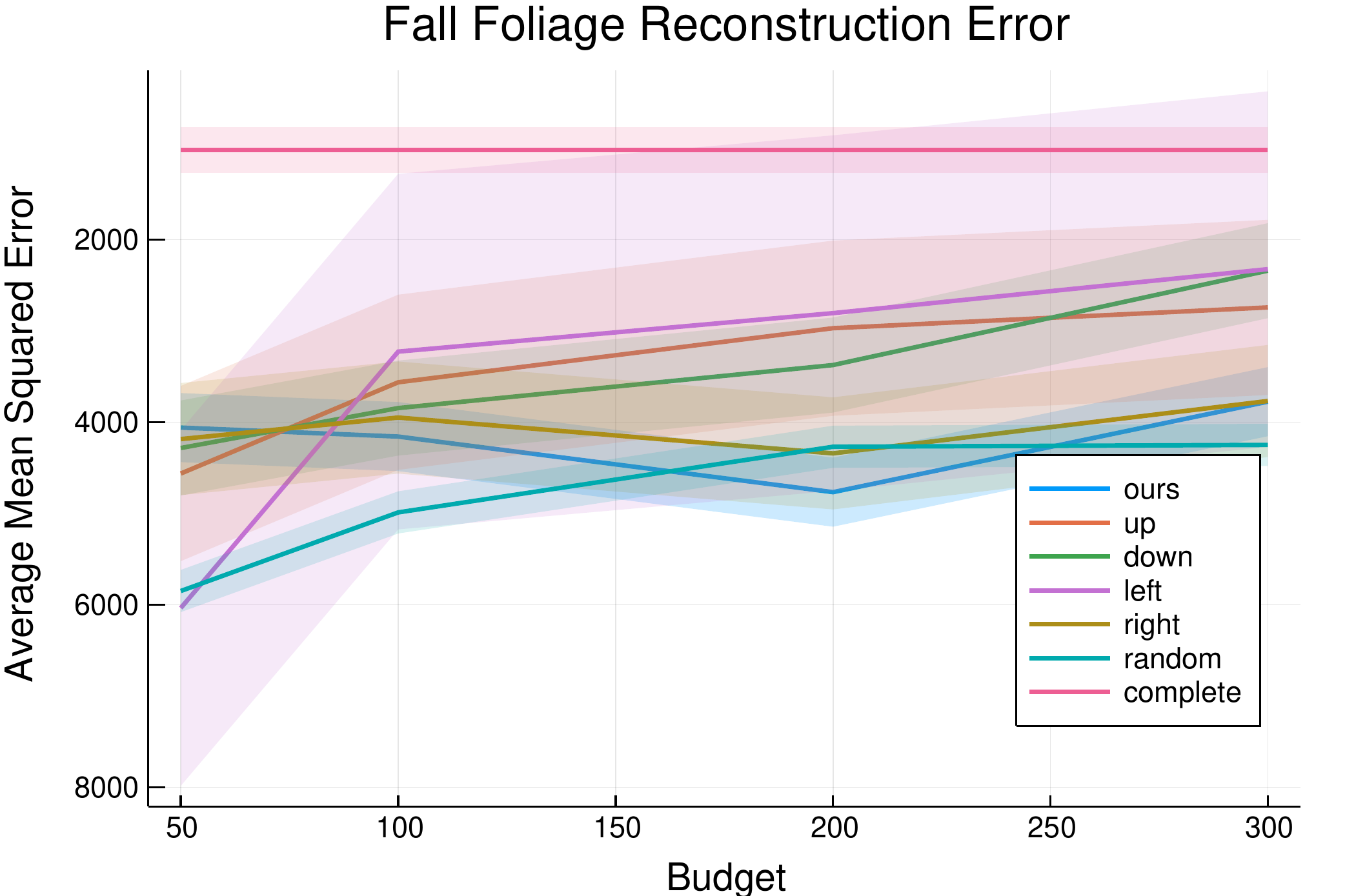}
            \caption{Reconstruction error for the Fall foliage dataset.}
            \label{fig:foliage_recon}
        \end{subfigure}
        \vskip\baselineskip
        \begin{subfigure}[b]{0.45\textwidth}   
            \centering 
            \includegraphics[width=\textwidth]{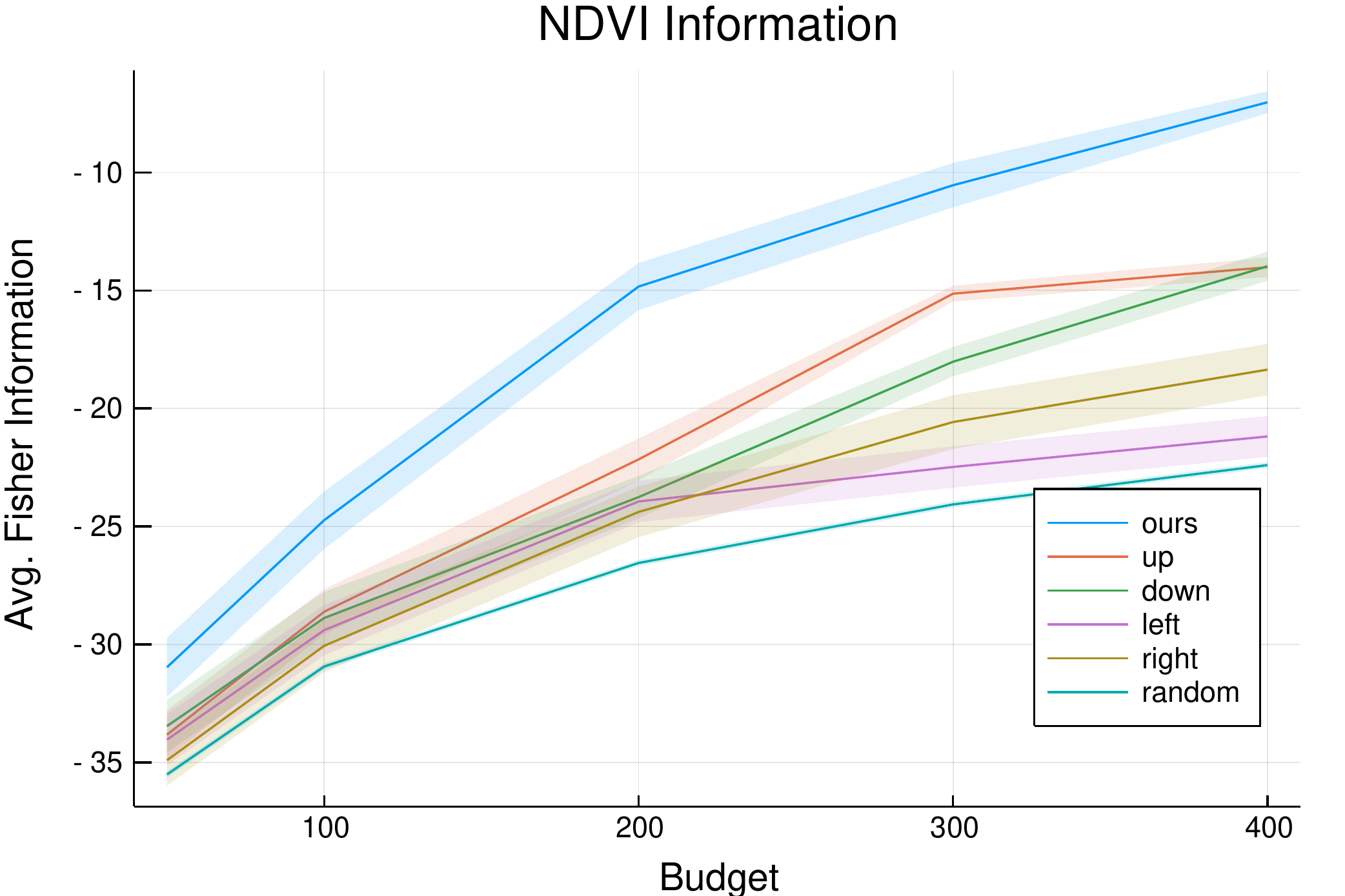}
           \caption{Fisher information for the NDVI dataset.}
            \label{fig:ndvi_info}
        \end{subfigure}
        \quad
        \begin{subfigure}[b]{0.45\textwidth}   
            \centering 
            \includegraphics[width=\textwidth]{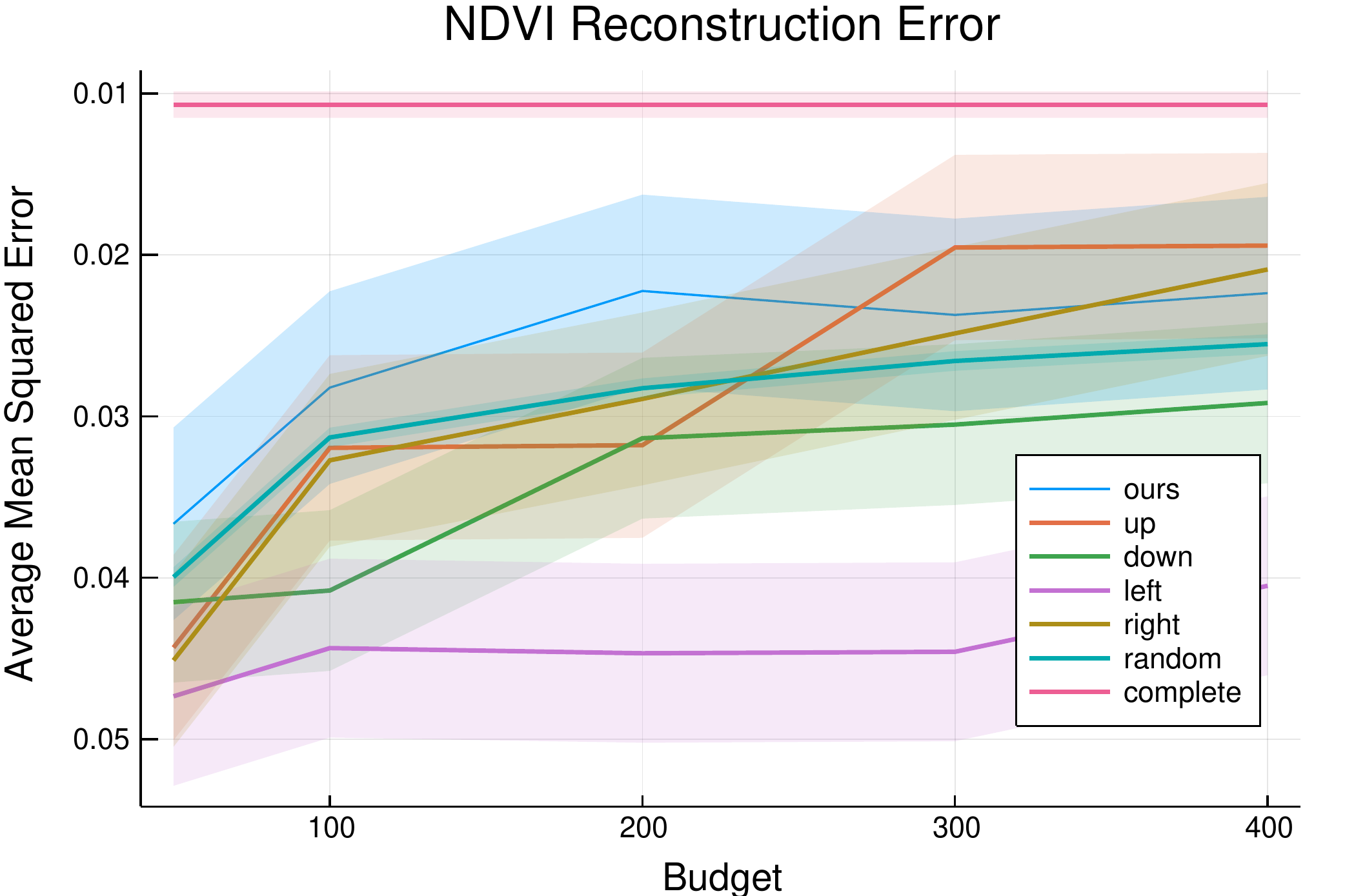}
             \caption{Reconstruction error for the NDVI dataset.}

            \label{fig:ndvi_recon}
        \end{subfigure}
        \caption{The average and standard error of Fisher information and mean squared reconstruction error. In each graph higher plots indicate higher performance.}
        \label{fig:graphs}
    \end{figure*}

% \begin{figure*}[t]
% \centering
% \includegraphics[width=0.2\textwidth]{figures/foliage_info_error.pdf}
%  \label{fig:foliage_info}
% \hfill
% \includegraphics[width=0.2\textwidth]{figures/foliage_recon_error.pdf}
%  \label{fig:foliage_recon}
% \hfill
% \includegraphics[width=0.2\textwidth]{figures/ndvi_info_error.pdf}
%  \label{fig:ndvi_info}
% \hfill
% \includegraphics[width=0.2\textwidth]{figures/ndvi_recon_error.pdf}
%  \label{fig:ndvi_recon}
%  \caption{The average and standard error of Fisher information and mean squared reconstruction error.}
% \label{fig:graphs}
% \end{figure*}

We evaluate our planning and prediction approach in simulation on two environment datasets: Fall Foliage and Normalized Difference Vegetation Index. For each dataset, a generalized low rank model is fit using an existing Julia implementation~\cite{udell2016generalized}. Each dataset is evaluated against four transect baselines and a random sampling baseline. The performance of the sampling methods are evaluated in terms of average Fisher information and average reconstruction error.

\subsection{Fall Foliage Data} \label{sec:foliage}

The Fall foliage dataset contains visual information about foliage in the northeast region of North America over the Fall season. Changes in foliage can be used to monitor changes in climate~\cite{xie2015green}. The attribute being monitored is the hue of the foliage, this value can range from $0$ to $360$.

The dataset is composed of 186 satellite images of the northeast region of North America, collected using the NASA Worldview tool.\footnote{We acknowledge the use of imagery from the NASA Worldview application (\texttt{worldview.earthdata.nasa.gov}), part of the NASA Earth Observing System Data and Information System (EOSDIS).} The images were captured in the RGB color space from September to November over the course of three years, 2013-2015. The original images were processed using hue values and canny edge detection \cite{scikit-image} to mask cloud cover and lighting artifacts. The processed dataset is composed of $298,$ $1076$ by $1629$ pixel images, all containing missing data. In order to monitor the appearance of the foliage, we covert the images to the HSV color space and only model and predict the hue of the images.

% In the northeast region of North America, among other places, seasonal temperature changes affect the appearance of foliage. Most notably, over the course of the Fall, most deciduous foliage changes from through green, yellow, orange, red, and brown -- approximately a monotonically-decreasing hue. The pace of this change varies year over year due to many factors including temperature and moisture. We are interested in tracking and predicting the evolution of the foliage's hue over the Fall season. This information can be used to determine the best times and locations to observe Fall foliage and for monitoring changes in climate.

\subsection{Normalized Difference Vegetation Index Data} \label{sec:ndviData}

The Normalized Difference Vegetation Index (NDVI) dataset contains NDVI values for the African continent. The NDVI is a measure of how much vegetation there is in a region. The value of the NDVI ranges from $-1$ to $1$, where areas with little to no vegetation are assigned low values and areas with high amounts of vegetation are assigned high values. A drop in NDVI can indicate a decline in vegetation health. Changes in the NDVI can therefore be used to predict and detect droughts~\cite{pettorelli2005using}. 

The NDVI dataset~\citep{ndvi_data} contains 310 grid-discretized daily NDVI maps collected by NOAA over the course of two years, 2013 and 2014. While the original dataset contains NDVI data for all of the continents, we only consider data from Africa. As a result our dataset consists of set of $1500$ by $1400$ grids of data. Ocean and lake regions do not have meaningful NDVI values, and are therefore not considered.

\begin{figure}[ht]
\centering
\includegraphics[width=\linewidth]{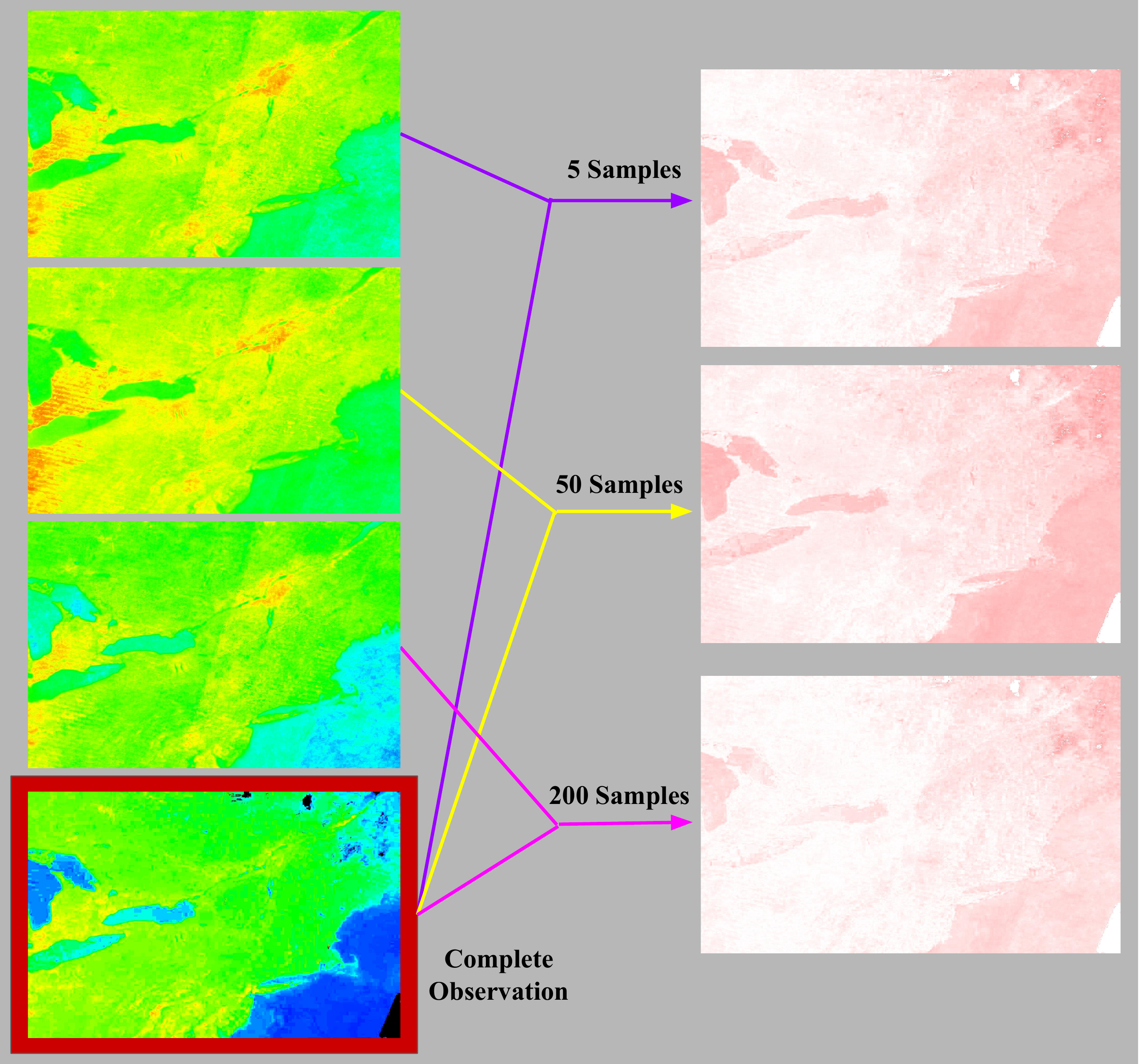}

\caption{The predicted states of the Fall foliage hue on September 15, 2015 achieved using our method and 5, 50 and 200 samples are shown in the right column. The images in the left column are error-heat maps. The darker a point in the heatmap, the greater the difference between the value predicted at that point and the actual value, as determined by the complete observation image.}
\label{fig:foliage}
\end{figure}

\begin{figure}[h]
\centering
\includegraphics[width=\linewidth]{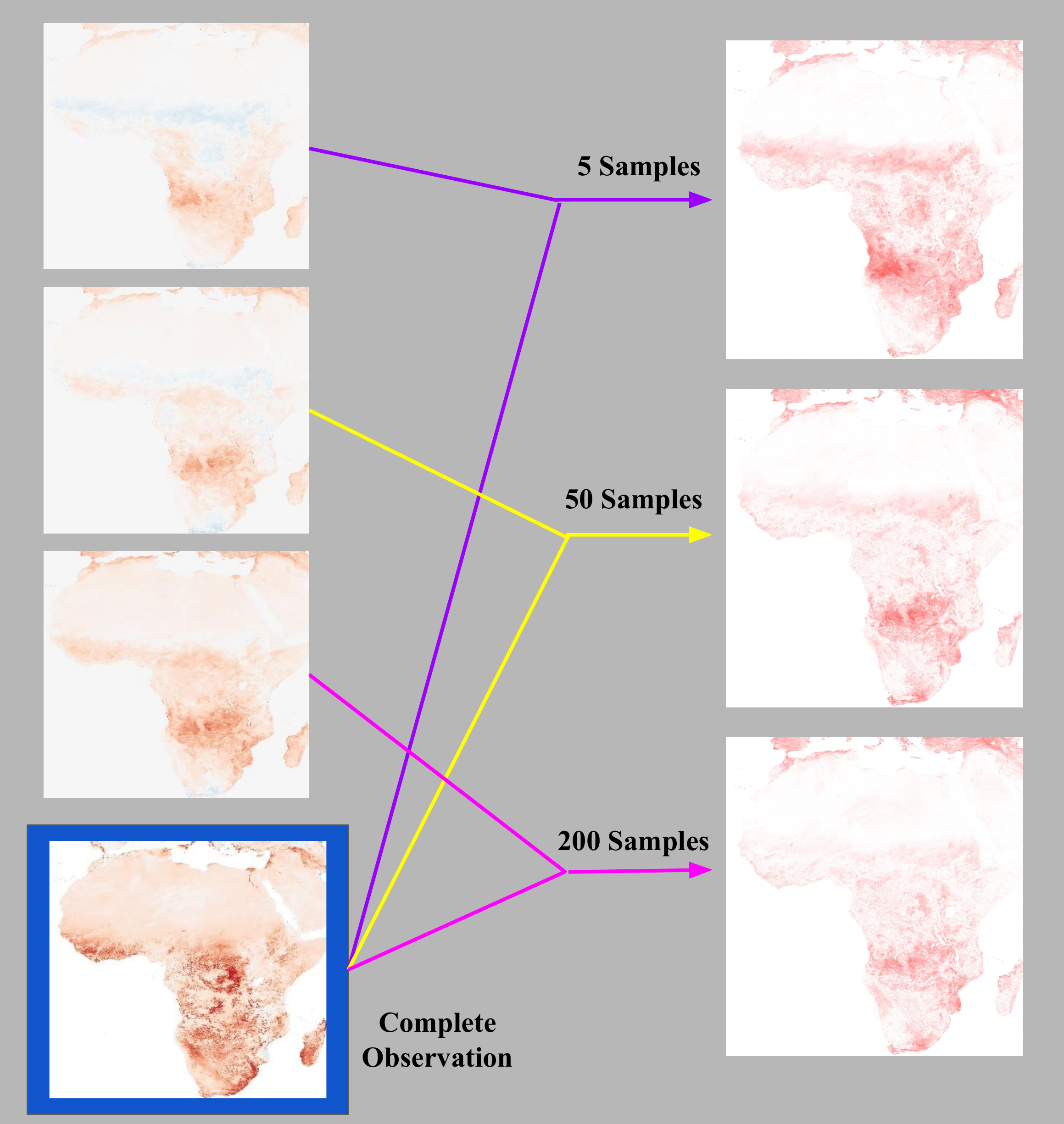}

\caption{The predicted states of the NDVI on March 29, 2014 achieved using our method and 5, 50 and 200 samples are shown in the right column. The images in the left column are error-heat maps. The darker a point in the heatmap, the greater the difference between the value predicted at that point and the actual value, as determined by the complete observation data. }
\label{fig:ndvi}
\end{figure}

\subsection{Baselines}\label{sec:baselines}

We compare our sampling method against transect, which is a common method for sampling in ecology~\cite{transect}, and random sampling baselines.

Transects are straight-line paths through a region. For our evaluation, we consider four directions of transect each with respect to the grid-discretization of the region: up, down, left, and right. For each direction of transect, the robot starts at the initial cell and moves in the given direction through the region, observing the state of \property{} at each cell it passes through. The length of the transect path is determined a Euclidean distance cost budget. In case that the path reaches a boundary of \region{} before the cost of the path equals the cost budget, the robot moves orthogonally for one step, and then moves in the opposite direction for the remainder of the budget.

The random-sampling baseline is a randomly-generated path through \region{} from a given initial cell. We assume every cell in \region{} has between three and eight neighboring cells. The random path is initialized to contain the initial cell, $P = [s]$. At each step of the process we compute the neighbors of the most recently added cell in $P$. If there exists a neighbor that is not in $P$, we randomly select one such cell and add it to $P$. If all of the neighbors are in $P$, we randomly select any neighbor and add it to $P$. This process is repeated until no cell can be added to $P$ while preserving the cost budget. The cost function is the ordered Euclidean distance between the cells in $P$.

\subsection{Results}

We use an existing Julia implementation~\cite{udell2016generalized} to fit a rank 5 GLRM to each dataset, using all but $9$ (Fall foliage) or $8$ (NDVI) held-out time snapshots. The held-out snapshots form the test sets used in our evaluation. 

We run each sampling method and on both datasets for ten distinct starting points and four (Fall foliage) or five (NDVI) different cost budgets, each run is referred to as a trial. Observations are taken from each snapshot in the test set at each sampling point generated by the trial. These observations are used to predict the state of all other points in the snapshot. For each dataset and cost budget, the average Fisher information is computed for the sampling sets produced by the corresponding trials. Reconstruction error is computed for each prediction. The reconstruction error is the sum of the mean squared error computed point-wise between the predicted states and the actual states in the corresponding test snapshot. The mean squared error is not computed for points missing in the test snapshot. The average reconstruction error is computed for each dataset and cost budget. One trial is performed per condition, with the exception of the random-sampler trials, which are performed 70 times per condition. 

The quality of the predictions is dependent on the number of points observed, and the missing data present in the Fall foliage dataset can result in very few observations from relatively large sampling sets. Therefore trials that resulted in fewer than 10 observations are excluded from the Fisher information and reconstruction error computations.

The results of the trials can be seen in \cref{fig:graphs}, where the shaded regions represent the standard error. The ``complete'' condition shown in \cref{fig:foliage_recon,fig:ndvi_recon} is the average mean squared error that occurs when predicting the state of each of the time snapshots in the test sets.
Qualitative reconstruction results are illustrated in \cref{fig:foliage,fig:ndvi}.

\subsection{Discussion}

Our planning method outperforms the baselines in terms of Fisher information, as can be seen in \cref{fig:foliage_info,fig:ndvi_info}. However, higher values of Fisher information do not necessarily result in more accurate reconstructions. \cref{fig:foliage_recon,fig:ndvi_recon} shows that the reconstruction error for our method is comparable to that of the baselines for both datasets and most trials. This is attributed to the fact that Fisher information is based on the expected value of the variables, and is not affected by the observed states. That is, points that are typically good predictors for the global state of the region may not hold as much information in the test snapshots; this could be due to missing or outlier data in the test set and the extent to which the model captures the structure of the data. Due to the inconsistency between Fisher information and reconstruction error, we conclude that Fisher information alone does not capture all information required to produce a good reconstruction.

% Our method tested on the NDVI dataset slightly outperforms the baseline for reconstruction error for smaller budgets, and has similar performance for higher budgets, as can be seen in \cref{fig:ndvi_recon}. Due to the inconsistency between Fisher information and reconstruction error, we conclude that Fisher information alone does not capture everything required to produce a good reconstruction. More research is needed to fully understand what properties of an observation lead to maximal reconstruction performance.

In order to visually interpret the reconstruction error plots we show the predictions made by our model, on one test time, after sampling from the sample set generated by our method with costs of $5,50$ and $200$. The raw reconstructions are shown in \cref{fig:foliage,fig:ndvi} as well as heatmaps representing the difference between the predictions and the original data. The value of the points in the heatmap images correlate to the reconstruction error. That is, darker points correspond to higher reconstruction error.

Since we have chosen to evaluate our method with rank $5$ models, at least $5$ samples are required to generate predictions. From the $5$ sample predictions we observe the structure contained in the models. That is, with only $5$ samples, the model is able to generate a reasonable prediction of the region that contains the basic elements that exist across the dataset. For example, a resonable outline of the land can be seen in the reconstructions for both datasets. In addition, in the general trend of lower values near the top of Africa can be seen. As more samples are added, the predictions on unseen points becomes more accurate as more sample specific features are defined. For example, the hue becomes more green in the Fall foliage prediction and the red regions of the NDVI prediction become darker. We note that the predictions generated by our model are far from perfect, we attribute this to the fact that only $200$ points are observed out of $1,752,804$ for the Fall foliage regions and $2,100,000$ for the NDVI regions.

\section{Conclusions, Limitations and Future Work}\label{sec:conclusion}
In this paper we have considered the problem of monitoring the state of a region, given a limited sampling budget and predicting the state of the unobserved portions. To this end, we have introduced a method for modelling the region and selecting a set of sampling points, such that the unobserved values can be predicted. We have shown that our path planning method outperforms the baselines in terms of information gain, but not in terms of reconstruction error. Our qualitative results illustrate that our method can be used to predict environmental features reasonably well with few observations. 

The GLRM used does not explicitly encode spatial or temporal correlations in the data, which may be limiting the accuracy of the predictions. That is, knowledge about the distance between two sampling points or the current time cannot be explicitly leveraged during the prediction process. Future work includes extending this method to incorporate spatiotemporal locality into the model.

Another limitation of this work is that it assumes the robot that can perfectly observe the state of each point in an environment and that both the robot and the environment exist in a discrete world. Furthermore, we discretize the map by assigning a single scalar value to a whole region in the physical environment.  For large regions, the problem of how to produce a single reading that accurately captures the overall quantity within the region becomes non-trivial. Furthermore, we cannot guarantee that a robot can travel between the given sampling points, as robot kinodynamics are not considered when choosing points. To make this method practical, future work includes generalizing our method to work in a continuous world, for robots with realistic kinodynamics and sensing abilities.

\section{Acknowledgment}

The authors would like to thank Matthew Luebbers for initial data collection and Robert Kleinberg for his helpful feedback.

\bibliographystyle{IEEEtran}
\bibliography{source}

% Generated by IEEEtran.bst, version: 1.14 (2015/08/26)
\begin{thebibliography}{10}
\providecommand{\url}[1]{#1}
\csname url@samestyle\endcsname
\providecommand{\newblock}{\relax}
\providecommand{\bibinfo}[2]{#2}
\providecommand{\BIBentrySTDinterwordspacing}{\spaceskip=0pt\relax}
\providecommand{\BIBentryALTinterwordstretchfactor}{4}
\providecommand{\BIBentryALTinterwordspacing}{\spaceskip=\fontdimen2\font plus
\BIBentryALTinterwordstretchfactor\fontdimen3\font minus
  \fontdimen4\font\relax}
\providecommand{\BIBforeignlanguage}[2]{{%
\expandafter\ifx\csname l@#1\endcsname\relax
\typeout{** WARNING: IEEEtran.bst: No hyphenation pattern has been}%
\typeout{** loaded for the language `#1'. Using the pattern for}%
\typeout{** the default language instead.}%
\else
\language=\csname l@#1\endcsname
\fi
#2}}
\providecommand{\BIBdecl}{\relax}
\BIBdecl

\bibitem{smith2011persistent2}
R.~N. Smith, M.~Schwager, S.~L. Smith, B.~H. Jones, D.~Rus, and G.~S. Sukhatme,
  ``Persistent ocean monitoring with underwater gliders: Adapting sampling
  resolution,'' \emph{Journal of Field Robotics}, vol.~28, no.~5, pp. 714--741,
  2011.

\bibitem{wang2010review}
J.~Wang, T.~W. Sammis, V.~P. Gutschick, M.~Gebremichael, S.~O. Dennis, and
  R.~E. Harrison, ``Review of satellite remote sensing use in forest health
  studies,'' \emph{The Open Geography Journal}, vol.~3, no.~1, 2010.

\bibitem{ma2016information}
K.-C. Ma, L.~Liu, and G.~S. Sukhatme, ``An information-driven and
  disturbance-aware planning method for long-term ocean monitoring,'' in
  \emph{2016 IEEE/RSJ International Conference on Intelligent Robots and
  Systems (IROS)}.\hskip 1em plus 0.5em minus 0.4em\relax IEEE, 2016, pp.
  2102--2108.

\bibitem{ma2018data}
K.-C. Ma, L.~Liu, H.~K. Heidarsson, and G.~S. Sukhatme, ``Data-driven learning
  and planning for environmental sampling,'' \emph{Journal of Field Robotics},
  vol.~35, no.~5, pp. 643--661, 2018.

\bibitem{langford1993unmanned}
J.~S. Langford and K.~A. Emanuel, ``An unmanned aircraft for dropwindsonde
  deployment and hurricane reconnaissance,'' \emph{Bulletin of the American
  Meteorological Society}, vol.~74, no.~3, pp. 367--376, 1993.

\bibitem{shannon1948mathematical}
C.~E. Shannon, ``A mathematical theory of communication,'' \emph{Bell system
  technical journal}, vol.~27, no.~3, pp. 379--423, 1948.

\bibitem{smith2011persistent}
S.~L. Smith, M.~Schwager, and D.~Rus, ``Persistent monitoring of changing
  environments using a robot with limited range sensing,'' in \emph{2011 IEEE
  International Conference on Robotics and Automation}.\hskip 1em plus 0.5em
  minus 0.4em\relax IEEE, 2011, pp. 5448--5455.

\bibitem{yu2014correlated}
J.~Yu, M.~Schwager, and D.~Rus, ``Correlated orienteering problem and its
  application to informative path planning for persistent monitoring tasks,''
  in \emph{2014 IEEE/RSJ International Conference on Intelligent Robots and
  Systems}.\hskip 1em plus 0.5em minus 0.4em\relax IEEE, 2014, pp. 342--349.

\bibitem{charrow2015information}
B.~Charrow, G.~Kahn, S.~Patil, S.~Liu, K.~Goldberg, P.~Abbeel, N.~Michael, and
  V.~Kumar, ``Information-theoretic planning with trajectory optimization for
  dense 3d mapping.'' in \emph{Robotics: Science and Systems}, vol.~11.\hskip
  1em plus 0.5em minus 0.4em\relax Rome, 2015.

\bibitem{low2011active}
K.~H. Low, J.~M. Dolan, and P.~Khosla, ``Active markov information-theoretic
  path planning for robotic environmental sensing,'' in \emph{The 10th
  International Conference on Autonomous Agents and Multiagent Systems-Volume
  2}.\hskip 1em plus 0.5em minus 0.4em\relax International Foundation for
  Autonomous Agents and Multiagent Systems, 2011, pp. 753--760.

\bibitem{sharma2015greedy}
D.~Sharma, A.~Kapoor, and A.~Deshpande, ``On greedy maximization of entropy,''
  in \emph{International Conference on Machine Learning}, 2015, pp. 1330--1338.

\bibitem{hashemi2018randomized}
A.~Hashemi, M.~Ghasemi, H.~Vikalo, and U.~Topcu, ``A randomized greedy
  algorithm for near-optimal sensor scheduling in large-scale sensor
  networks,'' in \emph{2018 Annual American Control Conference (ACC)}.\hskip
  1em plus 0.5em minus 0.4em\relax IEEE, 2018, pp. 1027--1032.

\bibitem{nemhauser1978analysis}
G.~L. Nemhauser, L.~A. Wolsey, and M.~L. Fisher, ``An analysis of
  approximations for maximizing submodular set functions—i,''
  \emph{Mathematical programming}, vol.~14, no.~1, pp. 265--294, 1978.

\bibitem{ko1995exact}
C.-W. Ko, J.~Lee, and M.~Queyranne, ``An exact algorithm for maximum entropy
  sampling,'' \emph{Operations Research}, vol.~43, no.~4, pp. 684--691, 1995.

\bibitem{jorgensen2018team}
S.~Jorgensen, R.~H. Chen, M.~B. Milam, and M.~Pavone, ``The team surviving
  orienteers problem: routing teams of robots in uncertain environments with
  survival constraints,'' \emph{Autonomous Robots}, vol.~42, no.~4, pp.
  927--952, 2018.

\bibitem{nguyen2016real}
J.~L. Nguyen, N.~R. Lawrance, R.~Fitch, and S.~Sukkarieh, ``Real-time path
  planning for long-term information gathering with an aerial glider,''
  \emph{Autonomous Robots}, vol.~40, no.~6, pp. 1017--1039, 2016.

\bibitem{hollinger2014sampling}
G.~A. Hollinger and G.~S. Sukhatme, ``Sampling-based robotic information
  gathering algorithms,'' \emph{The International Journal of Robotics
  Research}, vol.~33, no.~9, pp. 1271--1287, 2014.

\bibitem{kwak2008path}
J.-y. Kwak and P.~Scerri, ``Path planning for autonomous information collecting
  vehicles,'' in \emph{2008 11th International Conference on Information
  Fusion}.\hskip 1em plus 0.5em minus 0.4em\relax IEEE, 2008, pp. 1--8.

\bibitem{zhang2009information}
G.~Zhang, S.~Ferrari, and M.~Qian, ``An information roadmap method for robotic
  sensor path planning,'' \emph{Journal of Intelligent and Robotic Systems},
  vol.~56, no. 1-2, pp. 69--98, 2009.

\bibitem{lu2014information}
W.~Lu, G.~Zhang, and S.~Ferrari, ``An information potential approach to
  integrated sensor path planning and control,'' \emph{IEEE Transactions on
  Robotics}, vol.~30, no.~4, pp. 919--934, 2014.

\bibitem{zhang2016submodular}
H.~Zhang and Y.~Vorobeychik, ``Submodular optimization with routing
  constraints,'' in \emph{Thirtieth AAAI Conference on Artificial
  Intelligence}, 2016.

\bibitem{frieden_2004}
B.~R. Frieden, \emph{Science from Fisher Information: A Unification}.\hskip 1em
  plus 0.5em minus 0.4em\relax Cambridge University Press, 2004.

\bibitem{leung2006planning}
C.~Leung, S.~Huang, N.~Kwok, and G.~Dissanayake, ``Planning under uncertainty
  using model predictive control for information gathering,'' \emph{Robotics
  and Autonomous Systems}, vol.~54, no.~11, pp. 898--910, 2006.

\bibitem{singh2009nonmyopic}
A.~Singh, A.~Krause, and W.~J. Kaiser, ``Nonmyopic adaptive informative path
  planning for multiple robots,'' in \emph{Twenty-First International Joint
  Conference on Artificial Intelligence}, 2009.

\bibitem{cai2009information}
C.~Cai and S.~Ferrari, ``Information-driven sensor path planning by approximate
  cell decomposition,'' \emph{IEEE Transactions on Systems, Man, and
  Cybernetics, Part B (Cybernetics)}, vol.~39, no.~3, pp. 672--689, 2009.

\bibitem{low2009information}
K.~H. Low, J.~M. Dolan, and P.~Khosla, ``Information-theoretic approach to
  efficient adaptive path planning for mobile robotic environmental sensing,''
  in \emph{Nineteenth International Conference on Automated Planning and
  Scheduling}, 2009.

\bibitem{yang2018oboe}
\BIBentryALTinterwordspacing
C.~Yang, Y.~Akimoto, D.~W. Kim, and M.~Udell, ``{OBOE}: Collaborative filtering
  for {AutoML} initialization,'' in \emph{ACM SIGKDD Conference on Knowledge
  Discovery and Data Mining (KDD)}, 2019. [Online]. Available:
  \url{https://www.kdd.org/kdd2019/accepted-papers/view/oboe-collaborative-filtering-for-automl-model-selection}
\BIBentrySTDinterwordspacing

\bibitem{udell2016generalized}
M.~Udell, C.~Horn, R.~Zadeh, S.~Boyd \emph{et~al.}, ``Generalized low rank
  models,'' \emph{Foundations and Trends{\textregistered} in Machine Learning},
  vol.~9, no.~1, pp. 1--118, 2016.

\bibitem{golden1980approximate}
B.~Golden, L.~Bodin, T.~Doyle, and W.~Stewart~Jr, ``Approximate traveling
  salesman algorithms,'' \emph{Operations research}, vol.~28, no. 3-part-ii,
  pp. 694--711, 1980.

\bibitem{xie2015green}
Y.~Xie, K.~F. Ahmed, J.~M. Allen, A.~M. Wilson, and J.~A. Silander, ``Green-up
  of deciduous forest communities of northeastern north america in response to
  climate variation and climate change,'' \emph{Landscape ecology}, vol.~30,
  no.~1, pp. 109--123, 2015.

\bibitem{scikit-image}
\BIBentryALTinterwordspacing
S.~van~der Walt, J.~L. {S}ch\"onberger, J.~{Nunez-Iglesias}, F.~{B}oulogne,
  J.~D. {W}arner, N.~{Y}ager, E.~{G}ouillart, T.~{Y}u, and the scikit-image
  contributors, ``scikit-image: image processing in {P}ython,'' \emph{PeerJ},
  vol.~2, p. e453, 6 2014. [Online]. Available:
  \url{https://doi.org/10.7717/peerj.453}
\BIBentrySTDinterwordspacing

\bibitem{pettorelli2005using}
N.~Pettorelli, J.~O. Vik, A.~Mysterud, J.-M. Gaillard, C.~J. Tucker, and N.~C.
  Stenseth, ``Using the satellite-derived ndvi to assess ecological responses
  to environmental change,'' \emph{Trends in ecology \& evolution}, vol.~20,
  no.~9, pp. 503--510, 2005.

\bibitem{ndvi_data}
E.~Vermote, {NOAA} {CDR} Program. (2019): {NOAA} Climate Data Record (CDR) of
  AVHRR Normalized Difference Vegetation Index (NDVI), Version 5. NOAA National
  Centers for Environmental Information. https://doi.org/10.7289/V5ZG6QH9.
  Accessed [July 2019].

\bibitem{transect}
R.~Díaz-Gamboa and J.~Navarro, \emph{Line Transect Sampling}.\hskip 1em plus
  0.5em minus 0.4em\relax CRC Press, 2014, pp. 47--61.

\end{thebibliography}

\end{document}